\documentclass[11pt, a4paper, logo, copyright, numbering]{googledeepmind}

\usepackage[utf8]{inputenc}

\usepackage[authoryear, sort&compress, round]{natbib}
\usepackage{microtype}
\usepackage{hyperref}
\usepackage{url}
\usepackage{booktabs}
\usepackage{lineno}
\usepackage{tikz}
\usetikzlibrary{positioning}

\usepackage{microtype}
\usepackage{hyperref}
\usepackage{url}
\usepackage{booktabs}
\usepackage{amsmath}
\usepackage{lineno}
\usepackage{graphicx}
\usepackage{sidecap}
\usepackage{caption}
\usepackage{wrapfig}
\usepackage{enumitem}
\usepackage{algorithm}
\usepackage{algpseudocode}
\usepackage{mdframed}
\usepackage{float}
\usepackage{placeins}

\usepackage[most]{tcolorbox}
\usepackage{listings}  
\usepackage{xcolor}    
\usepackage{sidecap, caption}

\usepackage{microtype}      
\usepackage{makecell}
\usepackage{mathtools}
\usepackage{wrapfig}
\usepackage{subfig}
\usepackage{amsmath}
\usepackage{amssymb}
\usepackage{mathtools}
\usepackage{amsthm}
\usepackage{tabularx}
\usepackage{colortbl}

\usepackage{wrapfig}
\usepackage{booktabs} 
\usepackage{ stmaryrd }
\usepackage{colortbl}
\usepackage{pifont}
\usepackage{enumitem}
\usepackage{algorithm}
\usepackage{vcell}
\usepackage{listings}  

\usepackage{microtype}
\usepackage{hyperref}
\usepackage{url}
\usepackage{booktabs}
\usepackage{amsmath}
\usepackage{lineno}
\usepackage{graphicx}
\usepackage{caption}
\usepackage{enumitem}
\usepackage{algorithm}
\usepackage{algpseudocode}
\usepackage{mdframed}
\usepackage{float}
\usepackage{placeins}

\definecolor{listinggray}{gray}{0.9} 
\definecolor{headergreen}{RGB}{52, 73, 94} %

\definecolor{darkblue}{rgb}{0, 0, 0.5}
\hypersetup{colorlinks=true, citecolor=darkblue, linkcolor=darkblue, urlcolor=darkblue}

\title{Multi-Turn Puzzles: Evaluating Interactive Reasoning and Strategic Dialogue in LLMs}

\correspondingauthor{kbadola@google.com and mehrankazemi@google.com.}

\author[1]{Kartikeya Badola}
\author[1]{Jonathan Simon}
\author[1]{Arian Hosseini}
\author[1]{\authorcr Sara Marie Mc Carthy}
\author[1]{Tsendsuren Munkhdalai}
\author[1]{Abhimanyu Goyal}
\author[1]{Tomáš Kočiský}
\author[1]{Shyam Upadhyay}
\author[2]{\authorcr Bahare Fatemi}
\author[1]{Mehran Kazemi}

\affil[1]{Google DeepMind}
\affil[2]{Google Research}

\begin{document}

\begin{abstract}
Large language models (LLMs) excel at solving problems with clear and complete statements, but often struggle with nuanced environments or interactive tasks which are common in most real-world scenarios. This highlights the critical need for developing LLMs that can effectively engage in logically consistent multi-turn dialogue, seek information and reason with incomplete data. To this end, we introduce a novel benchmark comprising a suite of multi-turn tasks each designed to test specific reasoning, interactive dialogue, and information-seeking abilities. These tasks have deterministic scoring mechanisms, thus eliminating the need for human intervention. Evaluating frontier models on our benchmark reveals significant headroom. Our analysis shows that most errors emerge from poor instruction following, reasoning failures, and poor planning. This benchmark provides valuable insights into the strengths and weaknesses of current LLMs in handling complex, interactive scenarios and offers a robust platform for future research aimed at improving these critical capabilities.
\\
\\
\textbf{Data available at \href{https://huggingface.co/datasets/arianhosseini/mt\_puzzles}{huggingface.co/datasets/arianhosseini/mt\_puzzles} }
\end{abstract}

\maketitle

\begin{figure}[h]
    \centering
    \includegraphics[width=\textwidth]{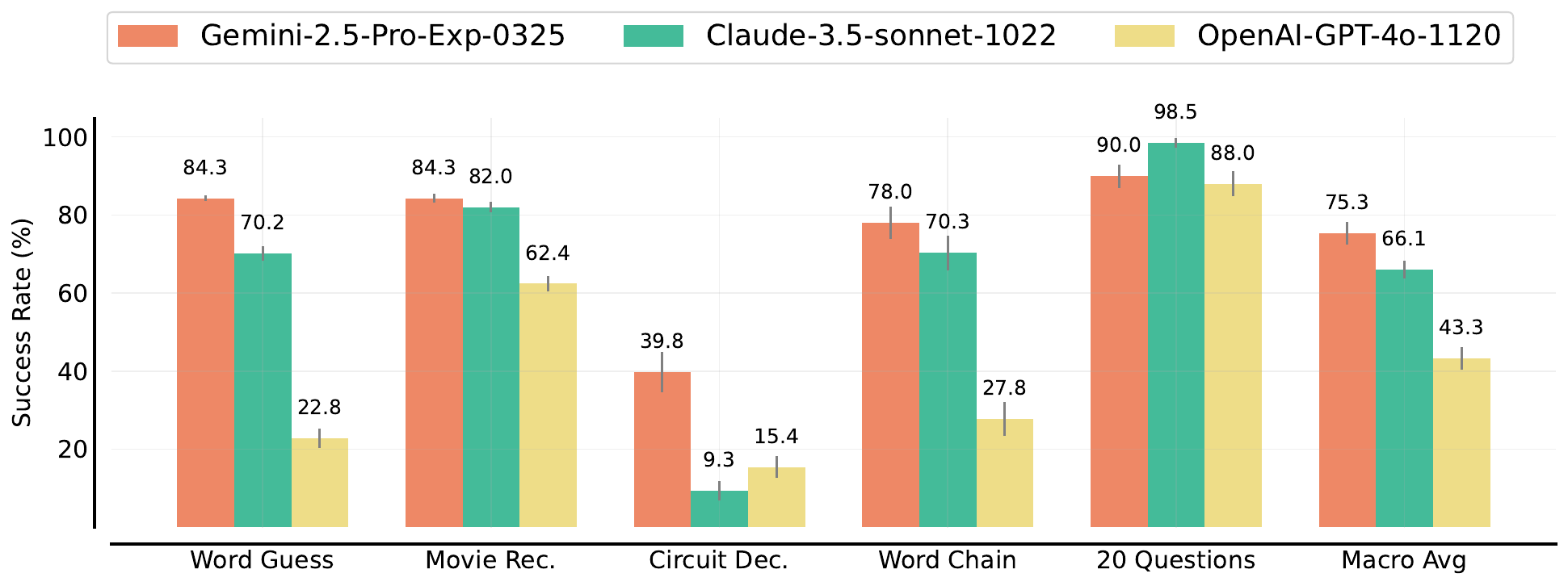}
    \caption{\textbf{Multi-Turn Puzzles (MTP) Benchmark:} performance of Gemini 2.5-Pro-Exp, Claude-3.5-sonnet and OpenAI-GPT-4o across five tasks. While models perform well on Twenty Questions, significant headroom remains for the rest of the tasks.}
    \label{fig:main_results}
\end{figure}

\vspace{-0.2cm}
\section{Introduction}

In situations where information is incomplete or hard to extract, large language model (LLM) agents must often take the initiative to proactively ask questions in a multi-turn interaction to gain more information, reduce uncertainty, and effectively solve problems.
This requires elaborate, interactive, strategic planning and reasoning to decide the next move in a multi-turn conversation as well as a robust memory of previous interactions. 
This capability is crucial for LLMs to operate effectively in complex, real-world environments where information is not always provided in a clear, single query such as virtual assistants and coding assistants~\citep{zheng2024lmsyschatm}.
While LLM agents excel at single-turn problems (e.g.,  mathematical and coding problems~\citep{openai2023gpt4, team2024gemini, llama3modelcard, deepseekai2025deepseekr1incentivizingreasoningcapability} involving a single prompt), they often perform poorly when a user's instructions are unclear or insufficient and further information must be obtained from the user before providing a useful response. This is a fundamental problem requiring further study and development of LLMs for how to reason with incomplete information, stay consistent and logical across turns, and demonstrate information seeking behaviour to plan and reason efficiently with finite interactions.

To assess multi-turn reasoning and instruction-following capabilities of LLMs, we propose multi-turn puzzles, a novel benchmark consisting of five distinct tasks, each with unique challenges and verifiable metrics~(\autoref{tab:tasks_and_metrics}):
\begin{itemize}[leftmargin=*,nosep,label={--}]
    \item \textbf{Word Guess} tests the efficiency of deductive reasoning and the ability of models to engage in homogeneous turn taking behavior.
    \item \textbf{Movie Recommendation} assesses the model's capacity for information seeking in imperfect information scenarios and learning user preferences through interactive questioning.
    \item \textbf{Circuit Decoding} tests logical deduction and the ability of the models to learn about the inner-workings of a new tool through multiple interactions of feeding an input and observing the output.
    \item \textbf{Word Chain} probes the model's ability to retain and deterministically condition on information obtained through multiple rounds of symmetric interaction.
    \item \textbf{Twenty Questions} measures coherence and logical consistency in dialogue. 
\end{itemize}
\begin{figure}[t]
    \centering
    \includegraphics[width=.8\linewidth]{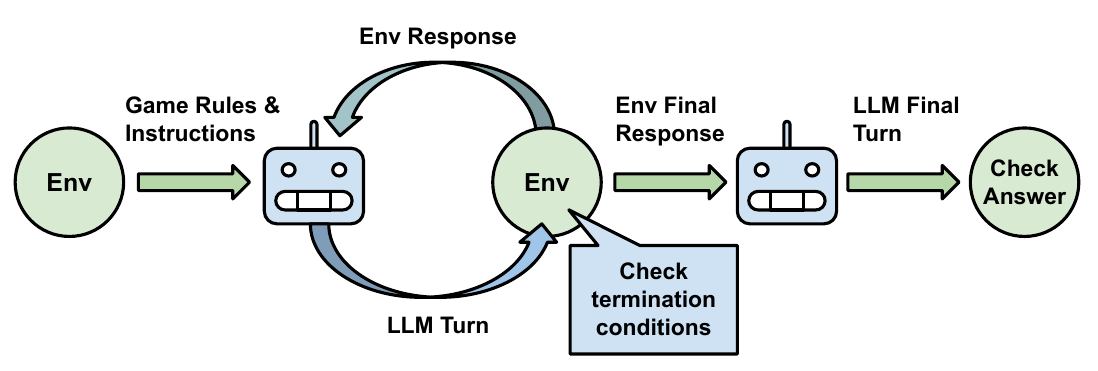}
    \caption{\textbf{Multi-Turn Interaction:} LLM interacts with the environment in the Multi-Turn Puzzles benchmark and the final output is checked using rule-based methods.}
    \label{fig:mt_env}
\end{figure}

Together these tasks provide a comprehensive evaluation of multi-turn reasoning and interactive problem-solving of LLMs, offering insights into how autonomous agents can be trained to behave more intelligently in ambiguous situations. 
We benchmark frontier models, including OpenAI GPT-4o, Anthropic Claude 3.5 Sonnet and Gemini 2.5 Pro Exp on our multi-turn tasks. \autoref{fig:main_results} indicates that while significant progress has been made, there remains a need for further exploration into the intricacies of multi-turn reasoning and information seeking behaviours. We believe this represents a valuable opportunity for the community to collaboratively advance the state of the art.

\begin{table*}[h]
\centering
\small
\renewcommand{\arraystretch}{1.1} 
\begin{tabular}{@{}m{1.7cm}|p{7.1cm}p{4.3cm}@{}}
\toprule
\textbf{Name} & \textbf{Short Description} & \textbf{Metric} \\
\hline
\vcell{Word Guess}
& \vcell{Guess the secret word in min attempts. Environment gives feedback on how close the guess is at each turn.\vspace{-8pt}}
& \vcell{Normalized number of attempts to solve the task.}\\
[-\rowheight]
\printcellmiddle
& \printcelltop
& \printcelltop\\\hline
\vcell{Movie Rec.}
& \vcell{Probe the user to decode the user preference function for N turns. Pick a movie for the same user at N+1 turn.\vspace{-6pt}}
& \vcell{Normalized rank of the final movie.} \\
[-\rowheight]
\printcellmiddle
& \printcelltop
& \printcelltop \\\hline
\vcell{Circuit Dec.}
& \vcell{Probe the C different boolean circuits for N turns. Predict the joint truth table of all the circuits at N+1 turn.\vspace{-6pt}}
& \vcell{ Normalized circuit-wise accuracy.} \\
[-1.002\rowheight]
\printcellmiddle
& \printcelltop
& \printcelltop \\\hline
\vcell{Word Chaining}
& \vcell{Model and user take turns choosing allow-listed words that start with the last letter of the previously chosen word. The episode ends successfully if no words with the correct starting letter remain, or max\_turns is hit.\vspace{-10pt}}
& \vcell{ Normalized \% of trajectories with successful end state.} \\
[-1.002\rowheight]
\printcellmiddle
& \printcelltop
& \printcelltop \\\hline
\vcell{Twenty Questions}
& \vcell{Model chooses a secret word. The user asks questions to determine what the word is. The model must answer these questions in a way that doesn't contradict previously-provided answers.\vspace{-10pt}}
& \vcell{ Normalized \% of logically consistent trajectories.} \\
[-1.002\rowheight]
\printcellmiddle
& \printcelltop
& \printcelltop\\
\bottomrule
\end{tabular}
\caption{Multi-turn Puzzle Task Descriptions and Metrics}
\label{tab:tasks_and_metrics}

\end{table*}








\vspace{-0.2cm}
\section{Multi-Turn Puzzles (MTP) Benchmark}
Navigating reasoning in domains where all the information is not readily present through multi-turn dialogue is a core capability of human intelligence, yet current benchmarks fail to adequately evaluate this ability. Most prevalent single-turn evaluations often fail to capture the complexities of real-world interactions, where problem-solving frequently involves iterative questioning, hypothesis refinement, and dynamic adaptation to evolving information.
We aim to create a benchmark that specifically targets this gap. MTP consists of five diverse tasks, including Word Guess, Movie Recommendation, Circuit Decoding, Word Chaining and Twenty Question. These tasks are designed to assess a model's ability to (1) reason with partial information, (2) be logically consistent across turns, and (3) exhibit information-seeking behavior.

\vspace{-0.3cm}
\subsection{Data Creation and Metrics}
\begin{figure}[ht]
    \centering
    \includegraphics[width=0.87\textwidth]{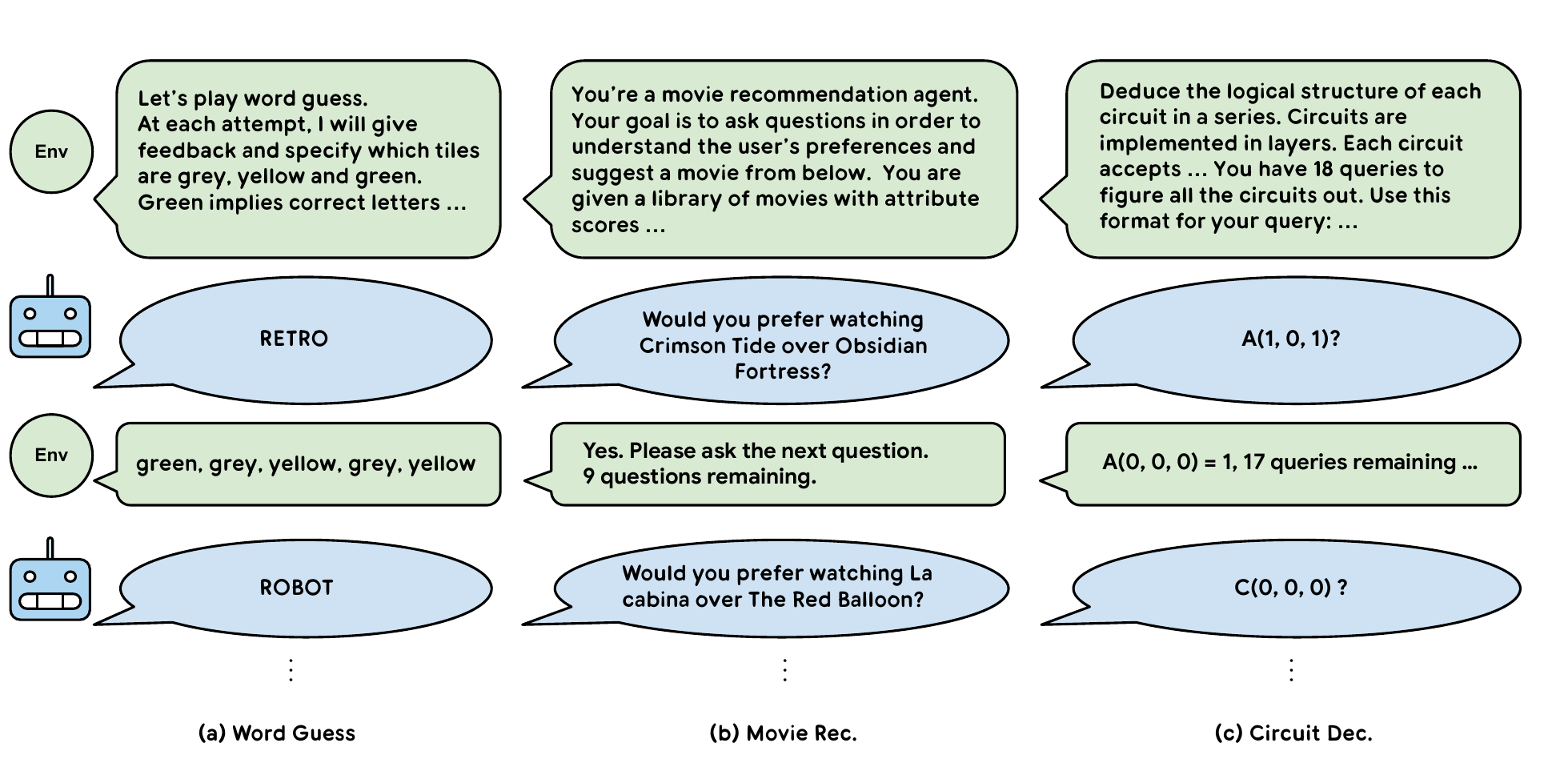}
    \includegraphics[width=0.65\textwidth]{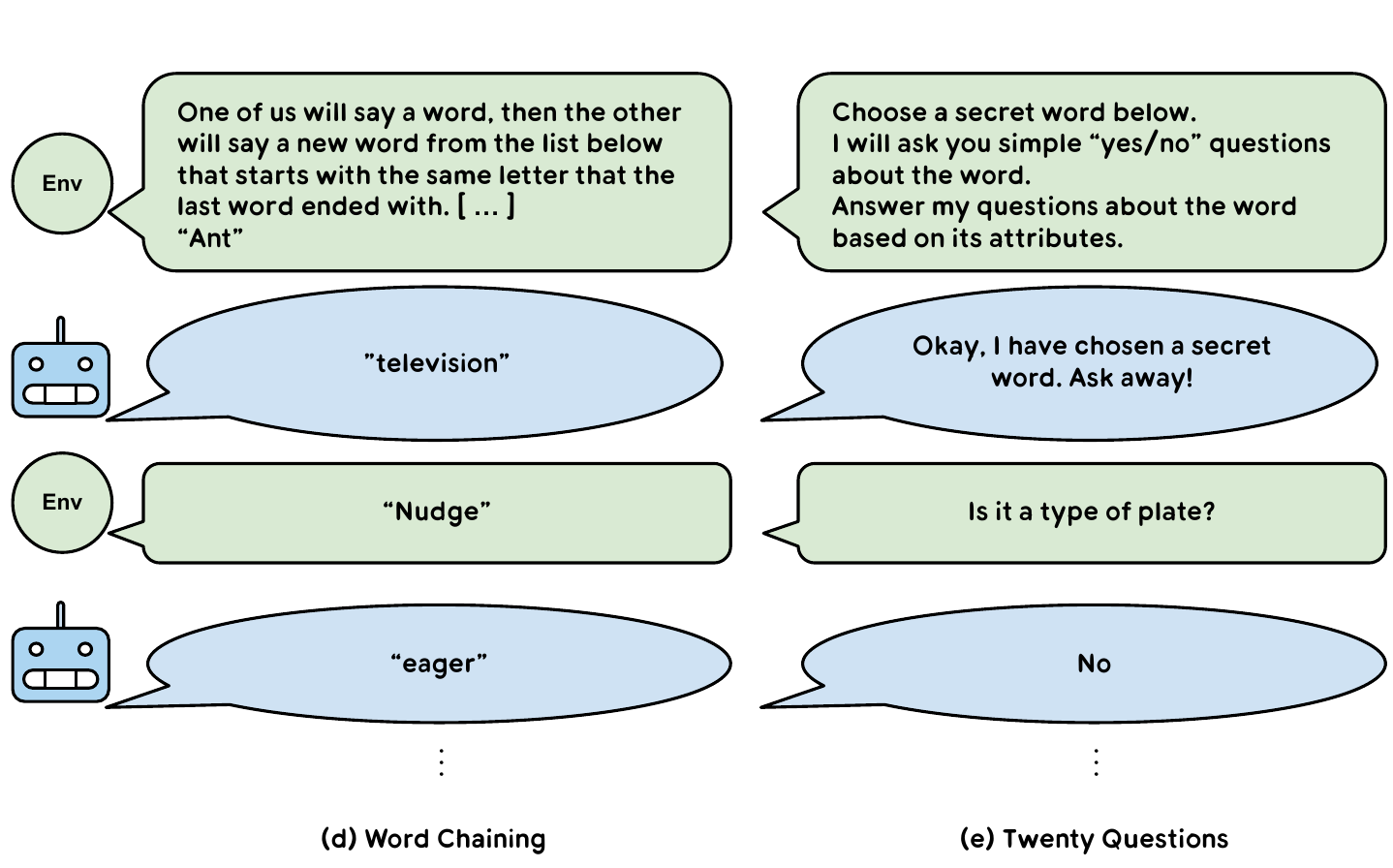}
    \caption{%
  \textbf{Examples of Multi-turn Puzzle Tasks.} (a) \textbf{Word Guess}: The model attempts to guess a secret word, receiving feedback on letter correctness and position after each guess. (b) \textbf{Movie Recommendation}: The model interactively asks questions to understand user preferences before suggesting a movie. (c) \textbf{Circuit Decoding}: The model queries boolean circuits with different inputs to deduce their logical structure. (d) \textbf{Word Chaining:} The model and environment take turns saying words from a list, where each new word must start with the last letter of the previous word. (e) \textbf{Twenty Questions:} The model selects a secret word and must answer the user's yes/no questions in a logically consistent manner.}
  \label{fig:samples}
\end{figure}

All of our tasks are synthetically created (see examples in~\autoref{fig:samples}). They utilize rule-based environments and deterministic scorers, eliminating the need for human-in-the-loop or autorators, and ensuring tight confidence intervals. \autoref{fig:mt_env} shows an overview of multi-turn interactions in our benchmark. This section presents a brief description of our task construction.

\vspace{-0.3cm}
\paragraph{Word Guess} The objective is to guess the secret word in the fewest attempts. We start with 10 unique vocabularies of 40 words. Each word has 5 letters. Since each word can serve as the secret word in a data point there are a total of 400 configurations. Section \ref{app:examples} outlines the full instruction for this task.

\vspace{-0.3cm}
\paragraph{Movie Recommendation} The goal is to discover the user's preference function over multiple turns and pick a movie for that user at the end. Each data point has a simulated user with a unique preference function, a seen set (for the questioning phase) and an unseen set (for the recommendation phase) of movies. There are 20 unique users, and 50 movie set variations, yielding a total of 1000 configurations for the task. Section~\ref{app:movie_rec_creation} describes how the users and the set of movies are initialized. Section~\ref{app:examples} outlines the instruction details for this task.

\vspace{-0.3cm}
\paragraph{Circuit Decoding} In this task, the objective is to determine the structure of a set of $C$ unknown boolean circuits. Each circuit receives $k$ boolean inputs and produces a single bit output. These $C$ circuits in total are made up of a fixed number of AND, OR and NOT gates. The model must deduce the functionality of all $C$ circuits, ultimately producing their combined joint truth table. For the evaluation set, we keep $C\times2^k=24$ and 300 unique circuit configurations are constructed. Further details are discussed in section~\ref{app:circuit_dec_rules}.


\vspace{-0.35cm}
\paragraph{Word Chaining} In each game, a lexicon is constructed by randomly sampling 500 words from SCOWL word list~\citep{atkinson_scowl}. The starting player is chosen randomly, and players alternate selecting words from the lexicon, adhering to: (i) the word must start with the previous word's final letter, and (ii) the word must not have been previously used. Rule violation leads to immediate loss. The game ends without loss if no valid moves remain or a turn limit is reached. The performance of an LLM is measured by its success rate (fraction of games completed without loss).

\paragraph{Twenty Questions} assesses an LLM's ability to maintain logical consistency during dialogue. The LLM selects a secret noun from a predefined WordNet lexicon~\citep{fellbaum1998wordnet} (80-100 words with provided hypernyms). It must then answer up to 20 binary (yes/no) questions posed by the user regarding the secret word or its hypernyms (including transitive ones). A loss occurs if and only if the LLM provides an answer that is logically inconsistent with its previous responses within the same game instance.

\subsection{Number of Turns and Episodes}
\autoref{tab:task_stats} shows the maximum number of turns allowed per episode for each task, alongside the average number of turns executed by Gemini 2.5 Pro Exp, Claude 3.5 Sonnet and GPT-4o during evaluation. Puzzles' maximum turns range from 11 (10 interaction turns plus 1 final prediction turn) for Movie Recommendation to 40 for Word Guess. Puzzles have early-stopping conditions which makes it possible for the game to terminate before the maximum number of turns has been reached. This can occur in cases such as the model making an irrecoverable error (e.g. logical contradiction in Twenty Questions), or when no valid playable moves remain (e.g. running out of playable words in Word Chaining).

\section{Experiments}
We report the performance of Gemini-2.5-Pro-Exp-0325, Claude-3.5-sonnet-1022 and OpenAI-GPT-4o-1120 on our benchmark. For each experiment, we use the default sampling parameters provided by each API.

\autoref{fig:main_results} illustrates the performance of these models across five different tasks. The results indicate that while models perform well on the \emph{Twenty Questions} task, there is significant room for improvement in the other tasks. To this end we probe model behavior through the lens of game duration, cost-dependent performance, inter-model turn taking, and information seeking strategies.

\begin{figure}[t]
    \centering
    \includegraphics[width=.9\textwidth]{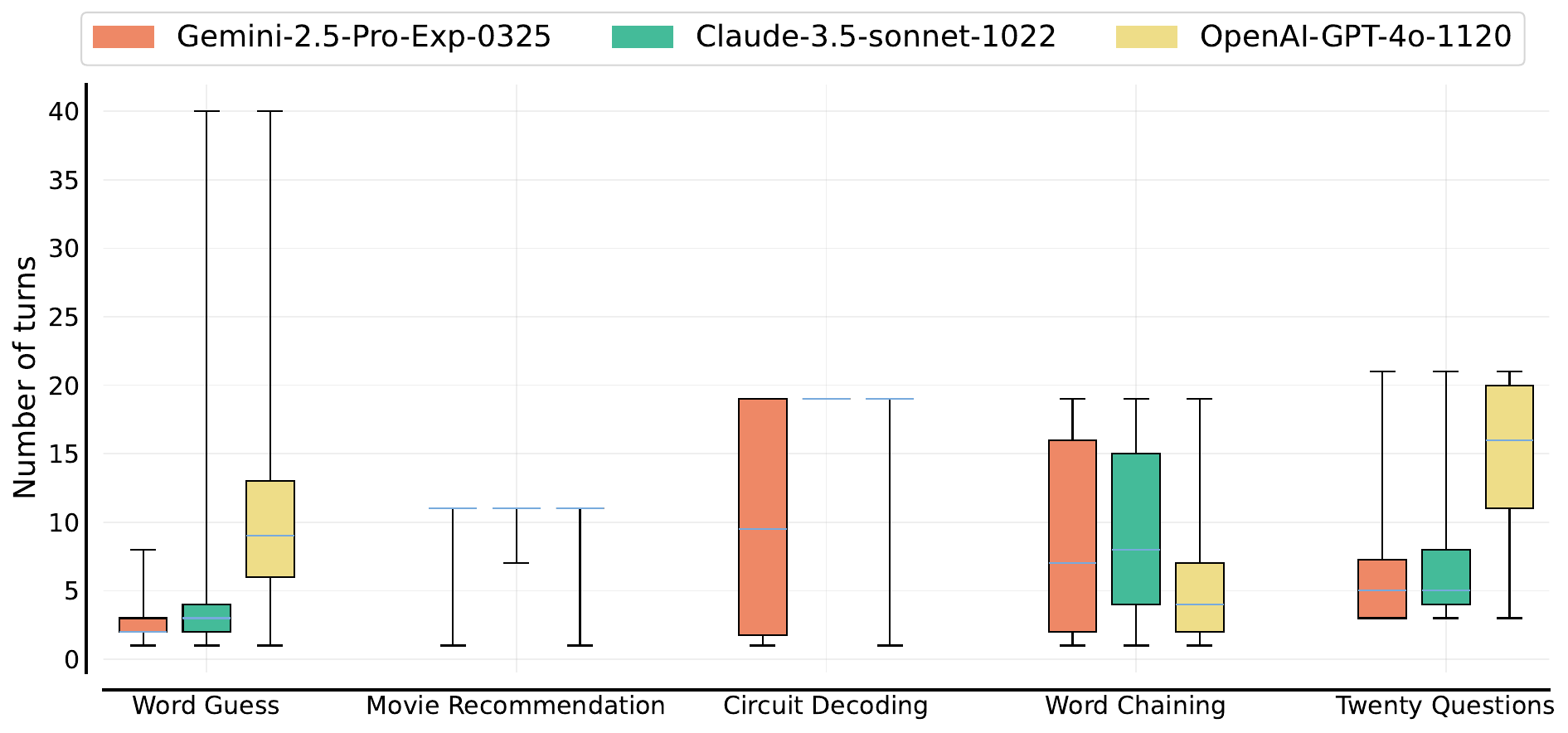}
    \caption{\textbf{Distribution of turns taken by models.} The same game type can have a different number of turns depending upon whether early termination conditions determining a win/lose were reached. Task constraints (such as fixed interaction lengths) and model specific strategies together shape number of conversation turns.}
 \label{fig:turn_analysis}
\end{figure}

\subsection{How Many Turns Do Models Take?}
\label{sec:num_of_turns}
Tasks such as Movie Recommendation, Circuit Decoding, and Twenty Questions have defined interaction limits followed by a final action, reflected in their maximum turn counts (e.g., 10+1, 18+1, 20+1 respectively). Notably, the average turns utilized by Gemini 2.5 indicate that some tasks (\autoref{tab:task_stats}), like Circuit Decoding (avg. 18.85 turns), consistently require nearly the full interaction budget, while others like Word Guess (avg. 5.37 turns) are often completed more quickly.

\autoref{fig:turn_analysis} shows the distribution of number of turns taken by each model across our tasks. In Movie Recommendation, the distribution for all models are tightly clustered around the maximum allowed 11 turns (10 interactions + 1 final), reflecting the task's design which necessitates utilizing the full interaction budget to gather user preferences before the final recommendation. Conversely, Word Guess shows significantly lower median turns across all models, particularly Gemini 2.5 and Claude 3.5, indicating that models often find the secret word efficiently, well before the generous 40-turn limit. Tasks like Circuit Decoding, Word Chaining, and Twenty Questions show greater variability. Circuit Decoding shows Claude 3.5 and GPT-4o consistently using the maximum 19 turns, while Gemini 2.5 displays a wider distribution, which in turn, is noticeable in their success rate on this task. Word Chaining shows considerable spread, reflecting difference in strategic depth. Different distributions in Twenty Questions show the duration for which logical consistency is maintained. Overall, the turn distributions effectively illustrate how task constraints (such as fixed interaction lengths) and model-specific strategies interact to determine conversation lengths.


\begin{figure}[t]
    \centering
    \includegraphics[width=\textwidth]{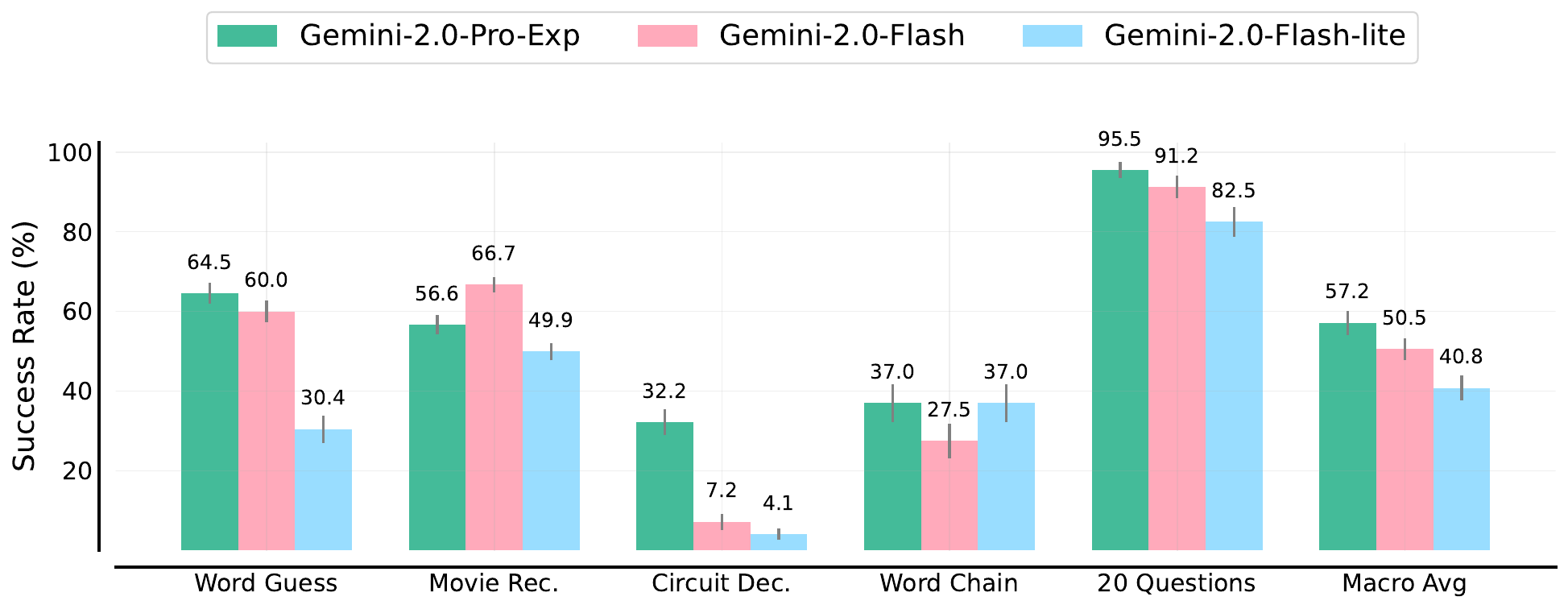}
    \caption{{Performance comparison of Gemini 2.0 Pro Exp, Gemini 2.0 Flash, and Gemini 2.0 Flash-lite models across the five multi-turn puzzle tasks.} The results show a clear cost-performance trade-off, with the larger Pro model achieving higher success rates than the more cost-efficient Flash and Flash-lite variants across all tasks and the macro average.}
 \label{fig:cost_vs_perf}
\end{figure}

\begin{table*}[t]
\centering
\small
\renewcommand{\arraystretch}{1.1} 
\begin{tabular}{@{}l|ccccc}
\toprule
\textbf{Task} & \textbf{Num Problems} & \textbf{Max Num Turns} & \textbf{Gemini 2.5} & \textbf{Claude 3.5} & \textbf{GPT-4o}\\
\hline
Word Guess & 400 & 40 & 2.4 & 3.8 & 12.9\\
Movie Rec. & 1000 & 11 (10 + 1) & 11 & 11 & 10.9\\
Circuit Dec. & 300 & 19 (18 + 1) & 10.5 & 19 & 18.4 \\
Word Chaining & 400 & 20 & 9 & 9.3 & 5.1\\
Twenty Questions & 400 & 21 (20 + 1) & 6.3 & 6.2 & 15.1 \\
\bottomrule
\end{tabular}
\caption{\textbf{Multi-Turn Puzzles statistics and average number of turns played by frontier models.} The games include early-stopping conditions making it possible for the game to end before the maximum number of turns has been reached. This can occur in cases such as the model committing a logical contradiction in Twenty Questions, or running out of playable words in Word Chaining.}
\label{tab:task_stats} 
\end{table*}

\subsection{Cost-Performance Trade-off}
\label{sec:cost_vs_perf}
Recent work suggests that cost-effective models reason systematically differently compared to their larger counterparts~\citep{hosseini2024llmreasonerscreatedequal}. Building on this, we extend our analysis to evaluate how these more affordable models perform specifically on our suite of multi-turn interactive tasks. We evaluated different variants within the Gemini 2.0 family – Pro-Exp, Flash, and Flash-lite – which represent distinct tiers in terms of computational cost, with Pro being the most expensive and Flash-lite the cheapest.

The results in~\autoref{fig:cost_vs_perf} reveal a clear trade-off between cost and performance. Gemini 2.0 Pro consistently achieved the highest success rates across all tasks, followed by Gemini Flash, and then Gemini Flash-lite. This trend is evident in the macro average scores. The performance gap varied depending on the task; for instance, the difference was particularly stark in Circuit Decoding (Pro: $32.2\%$, Flash: $7.2\%$, Lite: $4.1\%$), while relatively smaller, though still present, in tasks like Twenty Questions (Pro: $95.5\%$, Flash: $91.2\%$, Lite: $82.5\%$).

\subsection{Can Weaker and Stronger Models Finish Each Other's Thoughts?}
\label{sec:finish_thought}

Often a model might be prompted with an inaccurate or incomplete context of a dialogue or reasoning. It is important for the model to be able to detect flaws in the provided history, extract useful information, and pick up on the strategy which was used. In this regard, we investigated whether a stronger model (Gemini 2.5 Pro) could effectively conclude a conversation initiated by a weaker one (Gemini 2.0 Flash), and vice versa. This experiment involves one model handling the initial interactions ($n-1$), with the other model taking over for the final turn. We conduct this experiment on the Movie Recommendation task, which exhibits strong asymmetry between intermediate turns (preference probing) and the final turn (recommendation). 

When the weaker 2.0 Flash model finished the 2.5 Pro model's interactions, the score reached 75.2\%.
Surprisingly, when the stronger 2.5 Pro model completed the conversation started by the 2.0 Flash model, the success rate was similar at 75.7\%. Notably, substituting the weaker model for the final turn resulted in an 11\% performance drop (relative to the stronger model's baseline), whereas, allowing the stronger model to finish yielded a 13\% gain compared to the weaker' baseline. This underscores the impact of the final reasoning step in asymmetric tasks, indicating that a capable model executing the concluding turn can substantially mitigate deficiencies from earlier, less effective interaction turns.


\subsection{Qualitative Analysis}
\begin{figure}[t]
\begin{tikzpicture}[font=\tiny]
\definecolor{myDarkGreen}{RGB}{0, 100, 0}
  \node (d0) [] {Gemini 2.0 Flash-Lite};
  \node (d1) [below=0.1cm of d0, draw=blue, text width=6.2cm, rounded corners] {[Q1: Simply compares the first two movies]};
  \node (d2) [below=0.1cm of d1, text width=6.2cm, draw=blue, rounded corners] {Okay, based on the response, the user seems to appreciate movies that have more "Intellectual Depth", "Visual Details", "Realism Level" and "Emotional Intensity" and less "Pace", "Dialogue Focus", "Soundtrack Presence", "Character Complexity". [Many superficial conclusions] I will confirm my understanding with another question. [Q2: Compares two movies one high in realism and one high in dialogue focus]
  };
  \node (d3) [below=0.1cm of d2, text width=6.2cm, draw=blue, rounded corners] {
  Okay, the user clearly doesn't like "Pace" and "Dialogue Focus". Let's try to understand if the user cares about "Emotional Intensity" [wrong conclusion: the winner had a much higher dialogue focus.] [Q3: Compare two movies one high and one low in emotional intensity]
  };
  \node (d4) [below=0.1cm of d3, text width=6.2cm, draw=blue, rounded corners] {
  The user seems to like "Emotional Intensity". Now let's understand the importance of "Visual Details". [Q4: Compare two movies one high and one low in visual details]
  };
  \node (d5) [below=0.1cm of d4, text width=6.2cm, draw=blue, rounded corners] {
  [...] Now, I will try to evaluate how much the user cares about "Character Complexity". \\
  };
  \node (d6) [below=0.1cm of d5, text width=6.2cm, draw=blue, rounded corners] {
  [...] I am going to check how the user values "Soundtrack Presence". [Wrong probe: Compares two movies both high in soundtrack presence]
  };
  \node (d7) [below=0.1cm of d6, text width=6.2cm, draw=blue, rounded corners] {
  The user doesn't like "Soundtrack Presence" much [wrong conclusion]. Let's explore the importance of "Realism Level".
  };
  \node (d8) [below=0.1cm of d7, text width=6.2cm, draw=blue, rounded corners] {
  [...]
  };
  \node (d9) [below=0.1cm of d8, text width=6.2cm, draw=blue, rounded corners] {
  Based on the responses, the user seems to have the following preferences: [...] Now let's pick the best movie for the user from the provided dataset. [Q10: Asks a random question and the obtained information do not seem to get used.]
  };
  \node (e0) [right=5.2cm of d0] {Gemini 2.5 Pro};
  \node (e1) [below=0.1cm of e0, text width=7cm, draw=myDarkGreen, rounded corners] {
  [Q1: Compares two movies that are mostly similar, but one has high "Intellectual Depth" and one has high "Pace".]
  };
  \node (e2) [below=0.1cm of e1, text width=7cm, draw=myDarkGreen, rounded corners] {
  This gives me the first piece of information about your preferences.
  [Q2: Compares two movies that are mostly similar, but one has high "Intellectual Depth" and one has high "Pace" and high "Emotional Intensity".]
  };
  \node (e3) [below=0.1cm of e2, text width=7cm, draw=myDarkGreen, rounded corners] {
  [...] Your preference might lean towards Intellectual Depth over the other two [Pace and Emotional Intensity]. Let's test Emotional Intensity more directly.
  [Q3: Compares two movies, one has much higher "Emotional Intensity", the other has slightly higher values for several attributes.]
  };
  \node (e4) [below=0.1cm of e3, text width=7cm, draw=myDarkGreen, rounded corners] {
  [...] Both movies have the same Intellectual Depth and Pace. X has much higher Emotional Intensity, while Y has higher Dialogue Focus, Visual Details, and Realism Level. Your preference suggests that Emotional Intensity might be a significant positive factor for you, potentially outweighing Dialogue Focus, Visual Details and Realism. Let's test this further by comparing X against a movie with even higher Emotional Intensity but lower Pace.
  };
  \node (e5) [below=0.1cm of e4, text width=7cm, draw=myDarkGreen, rounded corners] {
  [...]
  };
  \node (e6) [below=0.1cm of e5, text width=7cm, draw=myDarkGreen, rounded corners] {
   [...] This might suggest that Intellectual Depth or Soundtrack Presence are quite important, or perhaps your preference regarding Pace or Emotional Intensity isn't as straightforward as previously thought. [Re-evaluating previous conclusions in face of new observations.] Let's try to isolate the effect of Soundtrack Presence versus Character Complexity and Realism Level.
  };
  \node (e7) [below=0.1cm of e6, text width=7cm, draw=myDarkGreen, rounded corners] {
   [...] 
  };
  \node (e8) [below=0.1cm of e7, text width=7cm, draw=myDarkGreen, rounded corners] {
   This is very informative. It strongly suggests that you highly value Character Complexity and Realism Level [...]. It also indicates that you likely dislike high Dialogue Focus, as "The Clockwork Mind" has a very high score in that attribute. This aligns well with previous observations about Character Complexity and Realism Level being important positive factors. For my final question, let's try to clarify the relative importance of Emotional Intensity versus Pace and Intellectual Depth.
  };

\end{tikzpicture}
\caption{\label{fig:qualitative} Sample (substantially simplified) model responses for Gemini 2.0 Flash-Lite (left) and Gemini 2.5 Pro (right) on Movie Recommendation showing their reasoning and planning process (for brevity, only model responses are shown). The text provided inside [] are extra information and context provided by the authors.}
\end{figure}
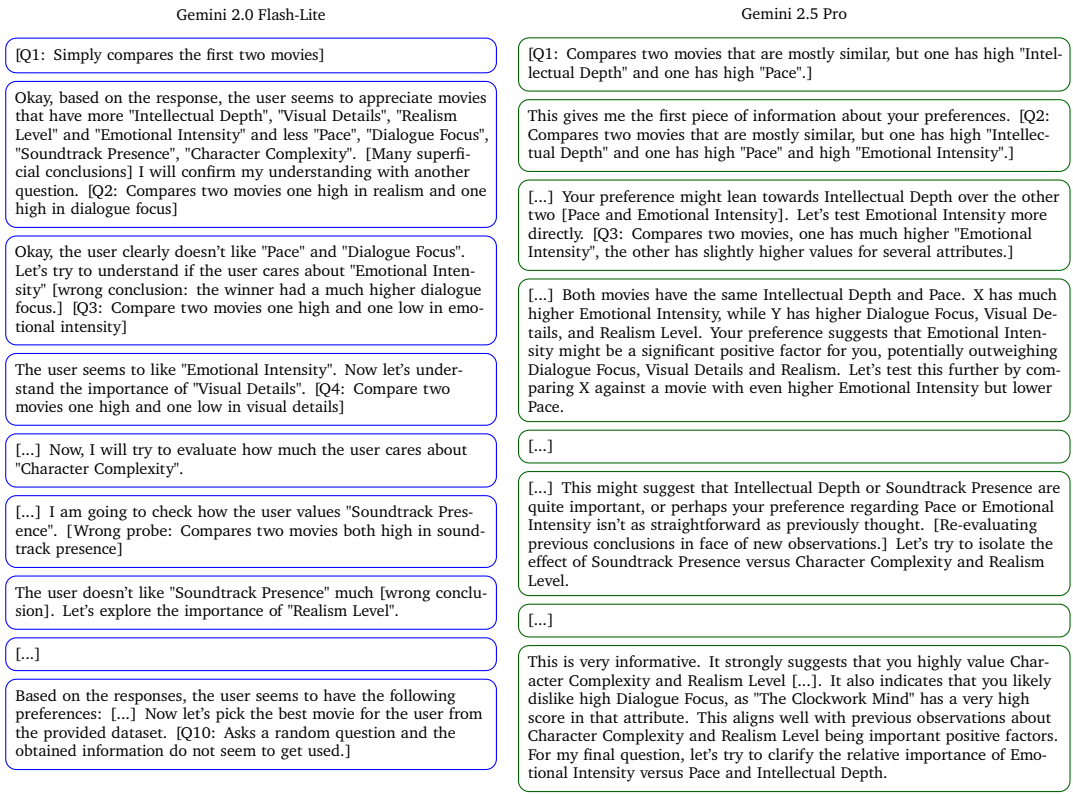

We conducted a qualitative analysis comparing a more capable model (Gemini 2.5 Pro) and a less capable model (Gemini 2.0 Flash-Lite) on our Movie Recommendation task. Our goal was to understand the differences in their reasoning and planning approaches and identify the main reasons for performance gaps between weaker and stronger models. Our analysis revealed the following key distinctions, with a sample dialogue from both models illustrated in Figure~\ref{fig:qualitative}.

\textbf{Reasoning Strategy:} The less capable model employs a simpler reasoning strategy, asking questions that focus on only one movie attribute at a time and drawing definitive conclusions about those attributes based on those individual responses. These questions often appear unrelated and do not contribute to a cohesive plan where subsequent questions build upon the information gained from previous ones. In contrast, the more capable model asks questions that consider multiple features simultaneously, allowing for richer insights from the user's responses. These questions are interconnected and form a more coherent plan, where the model designs its next questions based on observations from earlier interactions. This includes re-evaluating previous conclusions when new information contradicts them and investigating further. The samples provided in Figure~\ref{fig:qualitative} illustrate these differences in strategies.

\textbf{Asking Effective Questions:} When the models decide to seek specific information, we observed that the less capable model sometimes asks ineffective questions to obtain that information. For example, as shown in Figure~\ref{fig:qualitative}, the model might aim to understand the user's preference for "Soundtrack Presence" but then ask a question comparing two movies that both have a high level of this feature. Conversely, the more capable model generally compares movies that are similar in most other aspects and primarily differ in the specific feature(s) it wants to explore.

\textbf{Drawing Accurate Conclusions:} We found that some of the conclusions drawn by the less capable model are inaccurate or lack depth. For instance, in Figure~\ref{fig:qualitative}, we see that after the initial question, the model hastily makes numerous superficial assumptions about the user's likes and dislikes. Similarly, when comparing two movies that differ across multiple features, the model might draw conclusions about a single feature without considering the influence of the other differing features. However, the more capable model tends to derive more meaningful conclusions from the answers it receives and considers the interplay of various factors that could explain the user's feedback.

\textbf{Strategic Use of Questions:} Ideally, every question a model asks should contribute to gathering relevant information that aids in making a final recommendation. However, in the case of the less capable model, we observed instances where questions were not particularly useful. For example, the first question in Figure~\ref{fig:qualitative} simply compares the first two movies in the provided table, and the resulting information doesn't significantly contribute to the subsequent decision-making process. Additionally, the final question appears to be primarily asked to fulfill the task's requirements, as the model had already made a decision after nine questions. While definitively assessing the optimality of the more capable model's questions is challenging, we observed that it consistently probes for more relevant information that helps advance the task.

\section{Related Work}

\paragraph{Challenges in Multi-Turn Interaction.} The capabilities of LLMs have spurred significant research interest, initially focusing heavily on single-turn tasks. Recent studies have highlighted the challenges LLMs face when transitioning from single-turn success to effective multi-turn interaction. \cite{kwan-etal-2024-mt} attribute this drop in conversational tasks (like recollection and expansion) to factors like the distance to relevant context and error propagation.
\cite{liang2024mathchatbenchmarkingmathematicalreasoning} echo these findings specifically within the domain of mathematical problem-solving, noting a performance decline in interactive settings requiring sustained reasoning compared to single-turn math problems.
\cite{he2024multiifbenchmarkingllmsmultiturn} introduced the Multi-IF benchmark to assess multi-turn and multilingual instruction adherence, finding performance drops with each turn and particular difficulties with non-Latin scripts. 
Such studies underscore the need for better evaluation methodologies, hence we designed our tasks to only be solvable through logically and interactive dialogue.

\paragraph{Evaluating Multi-Turn Dialogue and Reasoning.} Various benchmarks aim to evaluate multi-turn capabilities. \cite{duan-etal-2024-botchat} evaluate the ability of LLMs to generate human-style multi-turn chat dialogues, using other powerful LLMs such as GPT-4~\citep{openai2023gpt4} as judges. Similarly, MT-Bench and Chatbot Arena both evaluate LLMs' multi-turn interaction quality~\citep{zheng2024judging}. MT-Bench uses a multi-turn question set and LLM-based ratings, while Chatbot Arena relies on pairwise comparisons from real user interactions.
\cite{zhang-etal-2024-probing} utilize an entity-deducing game where the model asks questions to probe conversational reasoning and planning. Our work complements this with tasks like Twenty Questions, where the model \textit{answers} questions, testing its logical consistency.
In contrast to these approaches focusing on general quality or instruction following, we evaluate LLMs specifically for multi-turn reasoning and information seeking abilities within puzzle scenarios using rule-based verifiable metrics.

\paragraph{Agentic Behavior and Interactive Environments.} Platforms such as LMRL Gym~\citep{abdulhai2023lmrlgymbenchmarksmultiturn} and AgentBench~\citep{liu2023agentbench} evaluate LLMs as agents in interactive settings. LMRL-Gym focuses on reinforcement learning (RL) capabilities (e.g. planning, credit assignment) in games like "guess my city", while AgentBench assesses instruction following, reasoning, and knowledge acquisition in simulated real-world environments like databases and web browsing~\citep{NEURIPS2022_82ad13ec, deng2023mind2webgeneralistagentweb}. In contrast, our benchmark provides simpler controllable game environments, allowing for more targeted evaluation of multi-turn capabilities. Additionally, many of the aforementioned games require an LLM to simulate the game environments and user interactions while we provide rule based environments.
\cite{zheng2025what} explores multi-turn prompting and fine-tuning for code generation, sharing our focus on multi-turn interaction. However, our puzzle environment uniquely requires logical deduction, strategic planning, and effective communication.
Other relevant work includes text-based game engines, such as TextWorld~\citep{cote18textworld} and games developed upon them, such as Coin Collector~\citep{xu2020deepreinforcementlearningstacked} and Jericho-QA~\citep{DBLP:journals/corr/abs-2006-07409}, primarily used for evaluating RL algorithms and agentic behavior. \cite{nie2024evolveevaluatingoptimizingllms, tajwar2025traininggenerallycuriousagent} explore in-context RL, focusing heavily on exploration in partially observable environments. \cite{hendrycks2021what} evaluate other LLM capabilities such as scientific reasoning and morality specifically in multi turn settings. 
For a comprehensive survey and further details regarding the multi-turn interaction capabilities of LLMs refer to \cite{zhang2025surveymultiturninteractioncapabilities}.
Unlike benchmarks relying on LLM judges or focusing broadly on agent capabilities in complex simulations, MTP uses simpler, controllable game environments with deterministic, rule-based scoring. This allows for targeted evaluation of core interactive reasoning skills, distinct from general conversational quality or instruction following alone.
\section{Conclusion}
While LLMs demonstrate strong performance on single-turn tasks, real-world applications often necessitate sustained interaction, multi-turn reasoning and dialogue abilities.
Our work introduces the Multi-Turn Puzzles (MTP) benchmark, contributing to this area by using interactive puzzle-solving as a specific testbed for these complex reasoning and dialogue planning skills. Evaluating current frontier models on MTP revealed significant performance disparities across tasks, highlighting substantial room for improvement, particularly in efficient information seeking (Movie Recommendation, Word Guess), complex deduction (Circuit Decoding), and strategic interaction (Word Chaining), although logical consistency (Twenty Questions) proved stronger.
Qualitative analysis further pinpointed key failure modes related to reasoning strategies, question formulation, conclusion accuracy, and planning. The MTP benchmark, with its rule-based environments and deterministic metrics, offers a robust platform for diagnosing these weaknesses. Future work should leverage such insights to further study how models can be fine-tuned for more effective multi-turn reasoning and coherent, strategic dialogue, ultimately enhancing their ability to navigate the complexities inherent in real-world interactions.
\bibliography{colm2025_conference}

\clearpage
\appendix




\section{Task Creation Details}

\subsection{Movie Recommendation}
\label{app:movie_rec_creation}
\begin{enumerate}
    \item \textbf{Users:}
    \begin{itemize}
        \item \textbf{User preference function}: We create a population of users, each having their own unique preference function modelled as a linear function of movie attributes:
        \[
        \text{score} = w_1 \cdot \text{attribute}_1 + w_2 \cdot \text{attribute}_2 + \dots + w_k \cdot \text{attribute}_k
        \]
        where $w_i$ are weights and $\text{attribute}_i$ are the movie attribute scores. Weights ($w_i$) are randomly generated for each user where $w_i \in [0, 1]$ and rounded to one decimal place
    \end{itemize}

    \item \textbf{Movies:}
    \begin{itemize}
        \item \textbf{Budget Constraint}:  To avoid outlier movies which would have high scores for all attributes, we assign a budget to each movie, limiting the total attribute scores:
         \[
         \text{movie\_budget} = \sum_{i=1}^{k} \text{attribute\_score}_i
         \]
         where $k$ is the number of attributes.  The $\text{movie\_budget}$ is randomly sampled between $(\text{base\_budget}, \text{base\_budget} + \text{budget\_delta})$.
        \item \textbf{Minimum Credits:} Each attribute is assigned a minimum score ($\text{attribute\_min\_credit}$).
        \item \textbf{Random Distribution:}  The remaining budget (after assigning minimum credits) is distributed randomly across attributes, ensuring diversity in movie profiles.
        \item \textbf{Seen vs. Unseen Sets:} Each episode has two sets of movies:
        \begin{itemize}
            \item \textbf{Seen Set:} Used for the agent's questioning phase, n=20
            \begin{itemize}
                \item Lower precision for attribute scores (fewer decimal places).
                \item  A pre-decided fraction of attributes may have their scores set to a $\text{min\_value}$. This makes it easier for the agent to compare movies that only differ in a few attributes.
            \end{itemize}
            \item \textbf{Unseen Set:} Used for final recommendation, n=40
            \begin{itemize}
                \item Higher precision for attribute scores.
                \item Doesn't have any sparsity like the seen set
            \end{itemize}
        \end{itemize}
    \end{itemize}
\end{enumerate}

\subsection{Circuit Decoding}
\label{app:circuit_dec_rules}
\begin{itemize}
    \item \textbf{Circuit Structure}: Each of the $C$ circuits has $k$ binary inputs and a single binary output.  Each circuit can have a depth greater than 1, meaning gates can be connected in multiple layers.
    \item \textbf{Constraint on logical operators}: All $C$ circuits, in aggregate, are composed of a fixed number of $x$ AND gates, $y$ OR gates, and $z$ NOT gates. This constraint on the *total* number of available gates is what makes sure that all circuits can be decoded without the need of brute force querying of each input combination.
    \item \textbf{Circuit Construction}: The available logic gates ($x$ AND, $y$ OR, $z$ NOT) are distributed among the $C$ circuits.  The distribution process ensures that all available gates are used and that each circuit has a sufficient number of AND/OR gates to combine the $k$ inputs into a single output. Specifically, each circuit must have at least $\lceil \log_2(k) \rceil$ AND + OR gates. Once the gates are distributed to a circuit, the circuit structure is generated layer by layer.  At each layer, any of the one operation can occur:
    \begin{itemize}
        \item Two input signals (outputs from the previous layer) are randomly selected and combined using either an AND or an OR gate (if available).
        \item Alternatively, a single input signal is randomly selected and inverted using a NOT gate (if available).
        \item All remaining input signals from the previous layer are carried forward to the next layer.
    \end{itemize}
    This process continues until all allocated gates for that circuit are used, resulting in a single output signal.
\end{itemize}

\subsection{Twenty Questions}
\label{app:twenty_questions}
\begin{itemize}
    \item \textbf{Game Progression}: The model is told to pick a secret word from a fixed set of words and their corresponding Wordnet hypernyms (i.e. attributes). The user then asks the model a series of yes/no questions about the attributes of the chosen secret word (e.g. "Is it a type of mammal?"), to which the model responds. If at any point the model responds in a way that logically contradicts the previously provided responses, the game immediately ends in a loss for the model. Otherwise the game continues until the user makes a final guess about the identity of the secret word (e.g. "Is your word chihuahua?") to which the model responds. The model's response to this final question is again assessed for logical consistency, with the game ending in a loss if it is inconsistent, and ending in a win otherwise. The user attempts to guess the final word when either (1) there is only a single logically consistent word remaining, (2) the maximum number of turns has been reached, or (3) a stochastic early-guessing condition occurs.
    \item \textbf{Logical Contradictions}: A logical contradiction can occur in the following ways:
    \begin{itemize}
        \item The user asks whether the secret word has a hypernym such that all words having that hypernym have already been excluded, and the model says "yes".
        \item The user asks whether the secret word has a hypernym such that all remaining logically consistent words have that hypernym, and the model says "no".
        \item There is one logically consistent word remaining, the user guesses it, and the model says "no".
        \item The user guesses a word that has already been logically excluded, and the model says "yes".
        \item No logically consistent words remain.
    \end{itemize}
    \item \textbf{User Actions}: The user simulator does not attempt to play optimally to efficiently identify the secret word. Instead, it operates based on a probabilistic state machine, that sometimes makes guesses to which the answer is already known, in an attempt to catch the model in a logical contradiction. On each turn, the user simulator selects one of the following actions with the given probability:
    \begin{itemize}
        \item Spontaneously attempt to guess the secret word, even if there are still multiple logically-consistent words remaining ($p=0.02$).
        \item When guessing the secret word, picking a word that has already been logically excluded ($p=0.15$).
        \item When guessing a hypernym, pick one that belongs only to words that have been logically excluded ($p=0.1$).
        \item When guessing a hypernym, pick one that belongs to all remaining logically consistent words ($p=0.1$).
        \item Guess a hypernym belonging to a proper subset of the non-excluded words (otherwise).
    \end{itemize}
    \item \textbf{Word Set}: The words and associated hypernyms used in the games are pre-filtered according to the following criteria:
    \begin{itemize}
        \item The words are nouns appearing in the SCOWL word list.
        \item Each word has at least 7 hypernyms.
        \item Every word has the hypernym "physical entity".
        \item No word has the hypernym "abstraction".
        \item No two words have identical sets of hypernyms.
        \item No word is the hypernym of any other word in the set.
    \end{itemize}
\end{itemize}

\clearpage

\subsection{Examples}
\label{app:examples}




\begin{tcolorbox}[colback=teal!6!white,colframe=teal!15!gray,
  colbacktitle=teal!12!white,title=\centering\color{black}\textbf{Movie Recommendation}]
  \footnotesize
\begin{lstlisting}[
    breaklines=true,
    postbreak={},
    breakindent=0pt,
    label={lst:movie},
    frame=none,
    numbers=none,              % Enable line numbers
    numbersep=2pt,            % Space between numbers and code
    xleftmargin=5pt,          % Left margin for code alignment
    stepnumber=1               % Show every line number
]
user
You're a movie recommendation agent who will interact with a user. Your goal is to ask questions in order to understand the user's preferences and suggest a movie that they'll enjoy the most. For this, you are given a library of movies in the table below. Each movie has scores on {num_attributes} different attributes: ...

The user's preference can be modeled as a linear function of the {num_attributes} scores and this function is hidden from you. The user will always act consistently based on the predefined user preference function.
...

You're only allowed to ask {num_questions} questions from the user, to uncover the underlying user preference function. Note that each question can only be formatted as follows:
"<question>Would you prefer watching $option_1 over $option_2?</question>".
...

The user will respond to your question. The response can only be "Yes", "No" or "No Preference". ...

Only once you're done interacting with the user after asking {num_questions} questions one-by-one, you will be given a new set of movies and based on whatever you've learnt about the user, your goal would be to pick the movie which you think the user will enjoy the most.
...

The movie dataset which you can use to ask the user questions is provided in the table below.
...

You can go ahead with asking the first question. While making deductions, try to uncover the underlying user preference function.

model
<question>Would you prefer watching Beneath the Crimson Tide over The Obsidian Fortress?</question>

user
Yes
Please ask the next question. 9 questions remaining. ...
...

user
No
Now apply whatever you learnt about the user and predict which movie will satisfy the user the most from the following set:
...

Reason about the linear user preference function and then provide your final answer. ...

model
<final_answer>A Dance with Time</final_answer>
\end{lstlisting}
\end{tcolorbox}

\begin{tcolorbox}[colback=teal!6!white,colframe=teal!15!gray,
  colbacktitle=teal!12!white,title=\centering\color{black}\textbf{Circuit Decoding}]
  \footnotesize
\begin{lstlisting}[
    breaklines=true,
    postbreak={},
    breakindent=0pt,
    label={lst:circuit_decoding},
    frame=none,
    numbers=none,              % Enable line numbers
    numbersep=2pt,            % Space between numbers and code
    xleftmargin=5pt,          % Left margin for code alignment
    stepnumber=1               % Show every line number
]
user
**Welcome to the Circuit Deduction Challenge!**

**Your Inventory:** You have a total of {components}. These are distributed across {num_circuits} circuits, labeled {circuit_labels}.

**Your Goal:** Deduce the logical structure of each circuit.

**The Setup:**
* Each circuit accepts {num_inputs} binary inputs (0 or 1) and produces a single binary output (0 or 1).
* Circuits are implemented in layers. Each layer takes two inputs from the previous layer and produces a single output.
* No circuit will always output 0 or always output 1, regardless of input.
* All circuit components are being used and are distributed without replacement across {num_circuits} circuits.

**Your Tools:** You have {num_questions} queries to figure all the circuits out. Each query is a call to a circuit with specific inputs.
At each turn of the conversation with me, you must make only 1 query. Hence, there will be {num_questions} turns of the conversation before you have to make your final guess.

**How to Play:**
1. At each turn, you must make two queries
2. Use the following format for your query:
...
3. You must make only 1 query at each turn

**Think Strategically:**
* Reason about the behavior of each circuit based on the available components.
* Plan your {num_questions} queries carefully to maximize information gained.

**Final Submission:**
After your {num_questions} queries, provide the truth tables for all circuits ({circuit_labels}) in this format:
...

Please start with your first query. Plan your steps before making the queries.

model
<query>A(0, 0, 0)</query><ctrl100>

user
A(0, 0, 0) = 1
Please send your next query. 17 queries remaining. ...
...

user
C(0, 0, 0) = 0
Please provide the truth tables for all circuits in the format requested. ...

model
<guess>10111111010001010101111111</guess>
\end{lstlisting}
\end{tcolorbox}

\begin{tcolorbox}[colback=teal!6!white,colframe=teal!15!gray,
  colbacktitle=teal!12!white,title=\centering\color{black}\textbf{Twenty Questions}]
  \footnotesize
\begin{lstlisting}[
    breaklines=true,
    postbreak={},
    breakindent=0pt,
    label={lst:twenty_questions},
    frame=none,
    numbers=none,              % Enable line numbers
    numbersep=2pt,            % Space between numbers and code
    xleftmargin=5pt,          % Left margin for code alignment
    stepnumber=1               % Show every line number
]
user
Let's play a game of 20 questions. I want you to choose a secret word (don't tell me what it is!) from the list below:

{word_list}

I will ask you simple yes/no questions about the word. You should answer these questions with either 'yes' or 'no', according to the attributes listed below:

{word_to_attributes_list}

So for example if your word is 'dog', and 'dog' has the listed attribute 'object', and I ask 'Is it a type of object?', you should say 'yes', even though dogs are not typically considered objects.
Conversely, if your word is 'dog', and 'dog' *does not* have the listed attribute 'pet', and I ask 'Is it a type of pet?', then you should say 'no', even though dogs are typically considered pets.

When I think I know what the word is, I'll ask you to confirm it, and you should again respond with either 'yes' or 'no'.

Ok, let me know once you've picked your word (remember, don't tell me what it is), and we'll get started!

model
Okay, I have chosen a secret word. Ask away!

user
Is it a type of plate?

model
no
...
\end{lstlisting}
\end{tcolorbox}

\begin{tcolorbox}[colback=teal!6!white,colframe=teal!15!gray,
  colbacktitle=teal!12!white,title=\centering\color{black}\textbf{Word Chaining}]
  \footnotesize
\begin{lstlisting}[
    breaklines=true,
    postbreak={},
    breakindent=0pt,
    label={lst:word_chaining},
    frame=none,
    numbers=none,              % Enable line numbers
    numbersep=2pt,            % Space between numbers and code
    xleftmargin=5pt,          % Left margin for code alignment
    stepnumber=1               % Show every line number
]
user
We're going to play a game. One of us will say a word, then the other will say a new word that starts with the same letter that the last word ended with. (So for example if I said 'ant', you might say 'television'.) All words must come from the list below. If you repeat a word, or select a word that isn't on the list, or break the first-letter rule, you lose. Please always wrap your chosen word in single-quote marks.

Allowed words: {words}

model
Okay, I'll start with 'encirclement'

user
Ok, my word is: 'topaz'.
...
\end{lstlisting}
\end{tcolorbox}

\begin{tcolorbox}[colback=teal!6!white,colframe=teal!15!gray,
  colbacktitle=teal!12!white,title=\centering\color{black}\textbf{Word Guess}]
  \footnotesize
\begin{lstlisting}[
    breaklines=true,
    postbreak={},
    breakindent=0pt,
    label={lst:word_guess},
    frame=none,
    numbers=none,              % Enable line numbers
    numbersep=2pt,            % Space between numbers and code
    xleftmargin=5pt,          % Left margin for code alignment
    stepnumber=1               % Show every line number
]
user
Let's play word guess. The secret word is from the following vocabulary: {vocab}.
At each attempt, I will give feedback to you and specify which tiles are grey, yellow and green. Green implies that the letter is in the word and in the correct position, yellow implies that the letter is in the word but in the wrong position and grey implies that the letter is not in the word.

For e.g., if the secret word is SPARK and you guessed PROOF, then the feedback would be:
yellow,yellow,grey,grey,grey.

Your goal is to correctly guess the secret word in the minimum number of attempts and will be evaluated on the number of attempts you take to arrive at the secret word. Think step by step to ensure that you're taking the most optimal actions to guess the secret word.

All of your guesses should be structured as follows:
<attempt>Your guess</attempt>

Make sure that all of your attempts are strictly from the vocabulary specified above.
Start with your first attempt and make sure to follow the format specified above.

model
<attempt>RETRO</attempt>

user
green,grey,grey,grey,yellow
...
\end{lstlisting}
\end{tcolorbox}



\end{document}